\documentclass[submission, creativecommons]{eptcs}
\providecommand{\event}{FMAS 2022} 

\usepackage{iftex}

  \usepackage{underscore}         
  \usepackage[T1]{fontenc}        

\usepackage[utf8]{inputenc}
\usepackage{amsmath}
\usepackage{amssymb}
\usepackage{amsfonts}
\usepackage{amsthm}
\usepackage{rotating}
\usepackage{xcolor}

\usepackage{wrapfig}
\usepackage{lscape}
\usepackage{epstopdf}

\usepackage{tikz}
\usepackage{tcolorbox}
\usepackage{todonotes}
\usepackage{booktabs}
\usepackage{tabularx}

\usepackage{tikz}
\usepackage{pgfplots}
\usepackage{pgfplotstable}
\pgfplotsset{compat=1.16}
\usepackage{subcaption}
\usepgfplotslibrary{groupplots}

\usepackage{graphicx}

\pgfplotsset{
SmallBarPlot/.style={
    font=\footnotesize,
    ybar,
    width=\linewidth,
    ymin=0,
    xtick=data,
    xticklabel style={text width=1.5cm, rotate=90, align=center}
},
BlueBars/.style={
    fill=blue!20, bar width=0.25
},
RedBars/.style={
    fill=red!20, bar width=0.25
}
}

\pgfplotsset{select coords between index/.style 2 args={
    x filter/.code={
        \ifnum\coordindex<#1\fi
        \ifnum\coordindex>#2\fi
    }
}}

\usetikzlibrary{patterns,arrows,arrows.meta,calc,shapes,shadows,decorations.pathmorphing,decorations.pathreplacing,automata,shapes.multipart,positioning,shapes.geometric,fit,circuits,trees,shapes.gates.logic.US,fit}

\tikzstyle{round} = [thick,draw=black,circle]
\newtheorem{example}{Example}
\theoremstyle{definition}
\newtheorem{definition}{Definition}[section]

\title{Scheduling for Urban Air Mobility using Safe Learning}
\author{Surya Murthy
\institute{School of Computer Engineering\\
University of Illinois, Urbana-Champaign \\
Urbana, Illinois}
\email{suryakm2@illinois.edu}
\and
Natasha A. Neogi 
\institute{NASA Langley Research Center\\ Hampton, Virginia}
\email{natasha.a.neogi@nasa.gov}
\and
Suda Bharadwaj
\institute{Skygrid LLC.\\ Austin, Texas}
\email{sbharadwaj@skygrid.com}
}

\def\titlerunning{Safe Learning for UAM Scheduling}
\def\authorrunning{S. Murthy, N. Neogi \& S. Bharadwaj}

\begin{document}
\maketitle

\begin{abstract}
{\bf Abstract.} This work considers the scheduling problem for Urban Air Mobility (UAM) vehicles travelling between origin-destination pairs with both hard and soft trip deadlines.  Each route is described by a discrete probability distribution over trip completion times (or delay) and over inter-arrival times of requests (or demand) for the route along with a fixed hard or soft deadline.  Soft deadlines carry a cost that is incurred when the deadline is missed.  An online, safe scheduler is developed that ensures that hard deadlines are never missed and that average cost of missing soft deadlines is minimized.  The system is modelled as a Markov Decision Process (MDP) and safe model based learning is used to find the  probabilistic distributions over route delays and demand.  Monte Carlo Tree Search (MCTS) Earliest Deadline First (EDF) is used to safely explore the learned models in an online fashion and develop a near-optimal non-preemptive scheduling policy.  These results are compared with Value Iteration (VI) and MCTS (Random) scheduling solutions. 
\end{abstract}

\section{Introduction}

The development of emerging aviation markets is crucial to maintaining international competitiveness and providing benefit to the US economy and general public. The safe development of an air transportation system that enables passengers, cargo, and payloads to execute their mission via aerial means has been a major driver in the past and current century.  
Urban Air Mobility (UAM) refers to the transportation of passengers and cargo over an urban environment, and it is often assumed to occur in an on-demand (aperiodic) manner.  UAM aims to alleviate urban congestion and improve urban public transit.  UAM vehicles are assumed to take off and land at a series of localized vertiports, vertihubs, or vertistations.  
Within the environment, the vehicles receive trip requests to move from one vertihub to another. By correctly scheduling these trip requests, the vehicles can navigate the UAM airspace safely and efficiently.  Thus, the development of a safe scheduler (i.e., no hard request misses its deadline and all hard requests are serviced) which does not require clairvoyance (i.e., scheduler does not have knowledge of the characteristics of the request, such as trip execution time or request inter-arrival time) is critical.
 A key question then becomes:  {\bf Given a set of origin-destination trip requests, where the requested time of arrival for each trip may either be a hard or soft deadline, can a schedule be created that guarantees all trips with hard deadlines arrive within their deadline and that the average cost of delayed trips with soft deadlines is minimized?}

\subsection{Contribution}
\label{subsec:PSC}
The major contributions of this work are:  (1) The infinite duration, aperiodic, non-preemptive scheduling problem for UAM trip requests with both hard and soft deadlines is formally modelled as a non-standard optimization problem over a Markov Decision Process (MDP).  (2)
The demand and delay distributions of the UAM trip requests (which are not known a priori) are safely learned over the MDP model using sampling techniques. (3) Monte Carlo Tree Search (MCTS) is used to learn safe, scalable non-preemptive scheduling strategies over the learned models. (4) Results are compared to solutions obtained using value iteration and Earliest Deadline First (EDF) scheduling techniques. The developed approach scales well for a small number of UAM trip routes where  each route can have infinitely many trip requests. 

\subsection{Related Work}
\label{subsec:RW}

The scheduling of UAM vehicles is a relatively new topic and has been considered from an operational concept point of view \cite{SWRMB2019}.  There have been some initial forays into building schedules for heterogeneous fleets of vehicles which focus on efficiency \cite{kim2019}.  Heuristic approaches for arrival sequencing have been put forth \cite{PP2018} as have energy efficient approaches \cite{KMP2018}.  A graph based reinforcement learning approach has been employed in \cite{PC2022}, where the UAM fleet scheduling problem is represented as an MDP  over a graph space, with the graph representing the network of vertiports (and their dynamic properties, e.g., demand). 
However, none of these approaches enforce safety guarantees over the synthesized schedule (i.e., all hard trip requests are completed on schedule).

The classical real-time scheduling problem has been explored in great detail \cite{dhall1978real},\cite{burns1991scheduling},\cite{buttazzo2012limited},\cite{davis2009survey},\cite{davis2011survey}. The scheduling of periodic tasks with hard real-time deadlines has been shown to be co-NP-complete \cite{LM1980},\cite{BHR1990}, and there has been considerable study of the case where the tasks are not strictly periodic \cite{HL1992},\cite{LSS1987}.  The scheduling of systems which have both hard and soft deadlines has also been investigated \cite{AMMM1999},\cite{LNL1987}. Classical algorithms for scheduling both hard and soft tasks in a preemptible fashion include Earliest Deadline First \cite{LL1973}, Earliest Deadline Late \cite{CC1989}, and Priority Exchange \cite{SB2004} among others.  These techniques do not consider probabilistic distributions across task completion times and inter-arrival times.

In \cite{GGR2018}, the stochastic scheduling problem with both hard and soft tasks on a single machine is considered, and a safe and efficient non-clairvoyant scheduler is computed in an online fashion.  The system is modelled as a finite MDP, and an algorithm for safe and optimal scheduler synthesis is developed.  In a follow-on effort by the authors contained in \cite{BCGPR2021}, safe learning for near-optimal scheduling is examined.  Monte Carlo Tree search techniques are leveraged to learn safe strategies for scheduling systems with large state spaces (e.g., $10^{20}$).  However, neither of these efforts consider non-preemptible tasks.   







\section {UAM Scheduling Problem}

The UAM scheduling problem is modeled as an MDP.  An MDP is defined as the tuple $(S, A, \delta, R)$ where: (1) $S$ is a finite set of states; (2) $A$ is a finite set of actions; (3) $\delta(s,a,s^\prime)$ is the probability that system goes from state $s$ to state $s^\prime$ after action $a$ is taken; and (4) $R$ is the reward obtained immediately after action $a$ is taken (with $|R| \leq B \in \mathbb{R}$ ).  Informally, the states of the MDP are the possible trip request configurations available within the system at the current time step.  The actions are the requests the scheduler can work on at that time step (i.e., the trips available to execute given that a UAM vehicle is available for that particular trip). The action space represents the actions the scheduler can take at any given state (i.e., the set of all possible trips that can be scheduled given UAM fleet availability).
\subsection{UAM Scheduling Term Definitions}
\label{subsec:UAMTerms}


\definition{\textbf{Trip completion time probability distribution ($\mathbf p$).}} Let $T$ be a discrete random variable drawn from a categorical distribution. The possible trip completion times are found in the support vector of the distribution denoted as $\mathcal{E}_T = \{0, 1, 2, ..., K\}$ where $K \in \mathbb{Z^{\text{nonneg}}}$ is the maximum possible trip completion time. Define the trip completion time probability distribution ${\mathbf p} = \{p_0,\dots, p_K\}$ where $p_i$ corresponds to the \emph{probability of the trip completion time $T$ being equal to i}.

\begin{example}
A trip completion time probability distribution of ${\mathbf p} = \{p_0 = 0, p_1 = 0, p_2 = 0, p_3 = 0.5, p_4 = 0.5\}$ has a maximum completion time $K = 4$ and a support vector of $\mathcal{E}_T = \{0, 1, 2, 3, 4\}$. Informally, this distribution means that the trip will complete in 3 time steps ($T = 3$) or 4 time steps ($T = 4$) with equal probability (0.5).
\end{example}

\definition{\textbf {Inter-arrival time probability distribution (q).} Define the inter-arrival time probability distribution $q = \{q_0,\dots, q_M\}$ with support vector $\mathcal{E}_V = \{0, 1, 2, ..., M\}$ where $M \in \mathbb{Z^{\text{nonneg}}}$ is the maximum possible inter-arrival time and the $i^{th}$ component of $q$  corresponds to the \emph{probability of the inter-arrival time $V$ being equal to i}.} 

\begin{example}
 An inter-arrival time distribution of $q = \{q_0 = 0, q_1 = 0,..., q_8 = 1\}$ has a maximum inter-arrival time $M = 8$, and a support vector of $\mathcal{E}_V = \{0, 1, 2,..., 8\}$. Informally, this distribution means that the a new trip request for that origin-destination pair will arrive in 8 time steps ($V = 8$) with probability 1.0.
\end{example}

\definition {\textbf{Deadline (D).}} Define $D$ as a positive integer that denotes the number of time steps by which a scheduled trip request \emph{must} be completed. 
In order to guarantee a trip can be scheduled without violating the deadline, assume $K \leq D \leq M$.

\begin{example}
A deadline of 7 is denoted by $D = 7$. Informally, this means that the trip request must be completed within 7 time steps of being started.
\end{example}

\definition{\textbf{Route.}} \label{def:route} A \emph{route} is a template for all trip requests of a given type. Define the set \textsc{Routes} as the set of all origin-destination vertiport pairs and denote its cardinality as $W$.

\definition{\textbf{Request.}}
A \emph{request} is a dynamic instantiation of a route. The following definition is presented for the case where there can only exist one request from a given route in the system at a time, without loss of generality. Formally,

\[r_i = (\mathbf{p_i}, D_i, \mathbf{q_i}) = (\{p_{i_{0}}, p_{i_{1}}, p_{i_{2}}, ..., p_{i_{K}}\}, D_i, \{q_{i_{0}}, q_{i_{1}}, ..., q_{i_{M}}\}): \forall \; i \in \textsc{Routes}.\] 
  If there is no ambiguity, the subscripts in the tuples $p_i$ and $q_i$ may be dropped (e.g., $r_i = (\{p_{0}, p_{1}, p_{2}, ..., p_{K}\}$,\\
  $D_i, \{q_{0}, q_{1}, ..., q_{M}\}$). Requests have an initial configuration. When a new instance of a request arrives, the request will be initialized in the initial configuration $r_{i_{in}} = (\mathbf{p_{i_{in}}}, D_{i_{in}}, \mathbf{q_{i_{in}}})$. Also, the initial trip completion time, initial deadline, and initial inter-arrival time cannot be equal to zero (e.g., $\mathbf{p_{i_{in}}},\mathbf{q_{i_{in}}} \neq \{1,0,..., 0\}, D_{i_{in}} \neq 0$). 

\begin{sidewaysfigure}
    \centering
    \includegraphics [width=19cm, height=15cm, keepaspectratio] {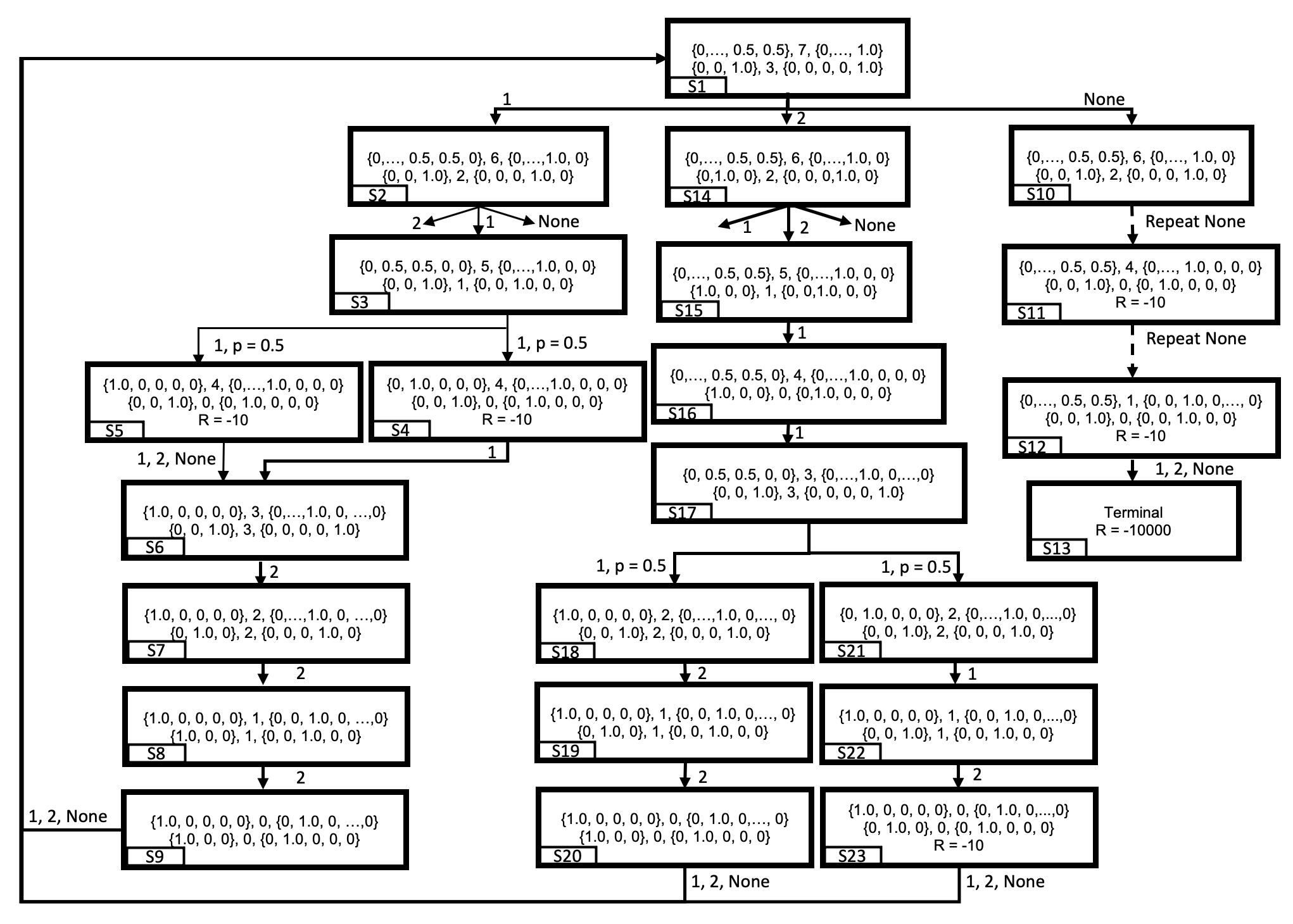}
    \caption{Markov Decision Process for Preemptive Scheduling Example}
    \label{fig:MDPExample}
\end{sidewaysfigure}

\begin{example}
Consider an MDP with 2 requests as seen in Figure \ref{fig:MDPExample}. The initial configuration of the first request is $( \{p_0 = 0, p_1 = 0, p_2 = 0, p_3 = 0.5, p_4 = 0.5\}, 7, \{q_0 = 0, q_1 = 0,..., q_8 = 1\})$. This request has a completion time distribution of $\{p_0 = 0, p_1 = 0, p_2 = 0, p_3 = 0.5, p_4 = 0.5\}$, a deadline of 7, and an inter-arrival time of $\{q_0 = 0, q_1 = 0,..., q_8 = 1\}$. Informally, it means that this request will take 3 or 4 time steps of dedicated work to complete, must be completed within 7 time steps, and will be replaced by a new request in 8 time steps. These values can change as time moves forward.
\end{example}

\definition {\textbf{Soft requests.}} The set of soft requests, $r_{soft}$, is defined as all of the requests that can be completed after their deadline reaches 0. However, a cost (or negative reward $R$) will be incurred for missing the deadline for that request. Formally,
$r_{soft} = \{(\mathbf p_0,D_0,\mathbf q_0), (\mathbf p_1,D_1,\mathbf q_1), ..., (\mathbf p_j,D_j,{\mathbf q}_j)\}$.

\begin{example}
In the MDP presented in Figure \ref{fig:MDPExample}, the soft request set contains the single request:  $( \{p_0 = 0, p_1 = 0, p_2 = 1\}, 3, \{q_0 = 0, q_1 = 0,... q_4 = 1\})$. This request can be completed even if 3 time steps pass, but a cost will be incurred.
\end{example}

\definition{{\textbf{Hard requests.}}}
The set of hard requests $r_{hard}$ are the requests that must be completed before the deadline reaches 0. Formally,
$r_{hard} = \{({\mathbf p}_{j + 1},D_{j + 1},{\mathbf q}_{j + 1}), ({\mathbf p}_{j + 2},D_{j + 2},{\mathbf q}_{j + 2}), ..., ({\mathbf p}_W,D_W,{\mathbf q}_W)\}$.

\begin{example}
In the MDP presented in Figure \ref{fig:MDPExample}, the hard request set contains the single request:  $( \{p_0 = 0, p_1 = 0, p_2 = 0, p_3 = 0.5, p_4 = 0.5\}, 7, \{q_0 = 0, q_1 = 0,... q_8 = 1\})$. This request must be completed within 7 time steps.
\end{example}

\subsection{MDP Construction}
\label{subsec:MDPC}
Define the UAM scheduling MDP $M = (S,A,R,\delta)$ as follows.


\definition{{\bf State Space S.}}
Define the state space as $S = \{r_{hard} \cup r_{soft}\} \cup \{s_{term}\}$ where there can exist only one request per route (without loss of generality), and $s_{term}$ is the \emph{terminal state} which is reached when a hard request misses its deadline. More formally,
$S =\bigcup_{i = 1}^{W} (\{p_0, p_1,..., p_K\}, D_i, \{ q_0, q_1,..., q_M\}) \cup s_{term}$.

\begin{example}
In the MDP presented in Figure \ref{fig:MDPExample}, the state space is comprised of the two requests: $( \{p_0 = 0, p_1 = 0, p_2 = 1\}, 3, \{q_0 = 0, q_1 = 0,... q_4 = 1\})$ and $( \{p_0 = 0, p_1 = 0, p_2 = 0, p_3 = 0.5, p_4 = 0.5\}, 7, \{q_0 = 0, q_1 = 0,... q_8 = 1\})$. The computation time, deadline and inter-arrival times of these requests decrement until they are completed and/or replaced by a new instance of the request (due to the evolution of time) resulting in the observed state space. In addition to the different configurations of the requests, the state space also contains a terminal state (S13).
\end{example}

\definition{{\bf Actions (A).}} Define an action $a_i \in A, i = {0, 1, 2, ..., W}$, which corresponds to the scheduling of the $i^{th}$ request $r_i$. Note that the $0^{th}$ action corresponds to idle time (e.g., no request is scheduled).  Formally, the set of actions $A$ is denoted $A = \{None, 1, 2, 3, ... , W\}$where $W$ is the maximum number of requests (i.e., there can be at most one active request for each route at a time, where the total number of routes is W, as seen in definition \ref{def:route}).

\begin{example}
In the MDP presented in Figure \ref{fig:MDPExample}, the set of actions is $A = \{None, 1, 2,..., W\}$. In state S1, a = 1 corresponds to working on $( \{p_0 = 0, p_1 = 0, p_2 = 0, p_3 = 0.5, p_4 = 0.5\}, 7, \{q_0 = 0, q_1 = 0,..., q_8 = 1\})$, and a = 2 corresponds to working on $( \{p_0 = 0, p_1 = 0, p_2 = 1\}, 3, \{q_0 = 0, q_1 = 0,..., q_4 = 1\})$. If $ a = None$, none of the requests are worked on and the scheduler remains idle.
\end{example}


\definition{{\bf Reward Function (R).}
Define the \emph{reward function} $R$ (where $R < 0$) as:
\begin{equation}
R(s,a,s') = 
    \begin{cases}
        wJ_{soft}, & \text{if } \forall \ r_i = ({p_0, ..., p_K}, D_i, {q_0, ..., q_M}) \in s'| r_i \in r_{soft} \wedge D_i = 0 \wedge p_{0} \neq 1 \\
        J_{hard}, & \text{if } s' =  s_{term}\\
        0, & \text{otherwise}
    \end{cases}
\end{equation}
}
where $J_{soft} < 0$ and $J_{hard} \ll J_{soft}$. 

The first case corresponds to the deadline of the soft request $r_i$ having elapsed ($D_i = 0$) and the request not having completed ($p_{0} \neq 0$), which incurs a small (negative) penalty of $J_{soft}$. There can be up to $w$ soft tasks whose deadlines elapse in a timestep, where $w \leq |r_{soft}|$. The second case corresponds to the deadline of the hard request $r_i$ having elapsed ($D_i = 0$) and the request not having completed ($p_{0_i} \neq 0$), which incurs a very large (negative) penalty of $J_{hard}$. Otherwise, no reward is incurred ($R(s,a,s') = 0$).

\begin{example}
As seen in Figure \ref{fig:MDPExample}, a soft task misses its deadline when transitioning from S3 to state S4 or S5. In both cases, the deadline for the soft request reaches 0 while the request is incomplete ($p_{0} \neq 1$). This results in a small negative reward of $J_{soft}=-10$ being accrued. A hard task misses its deadline in the transition from S12 to S13 in Figure \ref{fig:MDPExample}. In this transition, S13 is the terminal state, resulting in a large negative reward of $J_{hard}=-10000$ being accrued. 
\end{example}

\subsubsection{Transition Function} 

\label{subsubsec:Trans}
The construction of the transition function is presented as multiple individual cases that are complete when taken together. Formally, the \emph{transition function} $\delta: S\times A\times S^\prime \rightarrow [0,1]$ is defined as follows. 

{\bf Case 1: Agent idle ($\mathbf{a = \emph{None}}$) and no request is complete ($\mathbf{p_0 \neq 1}$), has a probability $p_1$ of completing, or has reached its deadline ($\mathbf{D \neq 0)}$.} 
In this case, the deadline and inter-arrival time of all requests will decrement while the completion times remain the same. Formally, $\delta(s,a,s') = 1$ if:
  \begin{eqnarray*}
  \forall \; r_i &=& (\{p_0, p_1, ..., p_K \}, D_i, \{q_0, q_1, ..., q_M \}) \in s \;|\; p_{1} = 0, D \neq 0, q_{1} = 0, \\ 
  a &=& None, \\
  r_i' &=& (\{p_0', p_1', ..., p_K' \}, D_i', \{q_0', q_1', ..., q_M' \})\; |\; \mathbf{p}'_i =  \mathbf{p}_i, D_i' = D_i - 1, q_{n}' = q_{{n+1}}, q_M' = 0, n = 0,..., M-1. 
  \end{eqnarray*}
  
\begin{example}
In the MDP in Figure \ref{fig:MDPExample}, this case occurs when transitioning from S1 to S10. When the agent chooses the action None, the computation time remains the same for both requests (e.g., for request 1, $\{p_0 = 0, p_1 = 0, p_2 = 0, p_3 = 0.5, p_4 = 0.5\} \rightarrow \{p_0' = 0, p_1' = 0, p_2' = 0, p_3' = 0.5, p_4' = 0.5\} $) while the deadline (e.g., for request 1, $ D_1 = 7 \rightarrow D_1' = 6 $) and inter-arrival time decrement (e.g., for request 1, $ \{q_0 = 0, q_1 = 0,..., q_8 = 1\} \rightarrow \{q_0' = 0, q_1' = 0,..., q_7' = 1, q_8' = 0\}$). 
\end{example}

{\bf Case 2: Agent works on request ${\mathbf{\ell}}$ and no request is complete, has a probability $p1$ of completing, or has reached its deadline.} When the agent chooses to work on the $\ell^{th}$ request ($a =  \ell \neq None$) and no request in the system is completed or has reached its deadline, 
the completion time, deadline, and inter-arrival time will decrement for the $\ell^{th}$ request,
and all other requests will decrement their  deadline and inter-arrival time while their completion time remains the same. Formally,  $\delta(s,a,s')$ = 1 if:
    \begin{eqnarray*}
        \forall r_i &=& (\{p_0, p_1, ..., p_K\},D_i, \{q_0, q_1, ..., q_M\}) \in s | p_{0} = 0, D_i \neq 0, q_{1} = 0,\\
    a &=& \ell,\\
     r_{\ell}' &=& (\{p'_0, p'_1,..., p'_K\}, \; D_{\ell}', \{q'_0, q'_1, ..., q'_M\})\; |\; p'_m = p_{m+1}, D_{\ell}' = D_{\ell} - 1, q_n' = q_{n+1} \; , p'_K = 0 , q_M' = 0\\
    & &\text{for }  m =0,..., K-1 \text{ and } n = 0,..., M-1,\\
    r_i' &=& ((\{p'_0, p'_1,..., p'_K\}, D'_i, \{q_0, q'_1, ..., q'_M\}), i \neq {\ell} |\; \mathbf{p}'_i = \mathbf{p}_{i}, D'_i = D_i - 1, q'_n = q_{n+1} \; , q'_M = 0 \\
    & &\text{for } n = 0,...,M-1. 
    \end{eqnarray*}

\begin{example}
In the MDP in Figure \ref{fig:MDPExample}, this case occurs when transitioning from S1 to S2. When the agent chooses action 1, the computation time decrements for the $1^{st}$ request ($ \{p_0 = 0, p_1 = 0, p_2 = 0, p_3 = 0.5, p_4 = 0.5\} \rightarrow \{p_0' = 0, p_1' = 0, p_2' = 0.5, p_3' = 0.5, p_4' = 0\} $) while the deadline (e.g., for request 1, $ D_1 = 7 \rightarrow D_1' = 6 $) and inter-arrival time ( e.g., for request 1, $ \{q_0 = 0, q_1 = 0,... q_8 = 1\} \rightarrow \{q_0' = 0, q_1' = 0,..., q_7' = 1, q_8' = 0\}$) decrement for both requests . 
\end{example} 
  
When the completion time for a request $r_{\ell}$ in the system equals 1 with a non-zero probability ($p_{\ell_{1}} \neq 0$) and the agent chooses to work on that request ($A = \ell$), the request will be completed with a probability $p_{\ell_{1}}$. On the other hand, the request will remain incomplete with probability $1 - p_{\ell_{1}}$. In this case, the non-zero probability entries of the next state will be increase by $p_{\ell_1} /{|p_{nz}-1|}$, where ${|p_{nz}-1|}$ is the remaining number of completion times with non-zero probability (i.e., if $p_{\ell_{m+1}} \neq 0$ then $p'_{\ell_{m}} = p_{\ell_{m+1}} + \frac{p_{\ell_1}}{ |p_{nz}-1|}$, otherwise $p'_{\ell_m} = p_{\ell_{m+1}}$).  The remaining requests all decrement their deadlines and inter-arrival times. Formally,$\delta(s,a,s') = 1 - p_{\ell_{1}}$ if:
  
\begin{eqnarray*}
    \exists r_i &=& (\{p_{i_{0}}, p_{i_{1}}, p_{i_{2}}, ..., p_{i_{K}}\}, D_i, \{q_{i_{0}}, q_{i_{1}}, ..., q_{i_{M}}\}) \in s | p_{i_{1}} \neq 0, D_i \neq 0, q_{i_{1}} = 0,\\
    a &=& \ell, \\
     r'_{\ell} &=& \left(\{p'_{\ell_{0}}, p'_{\ell_{1}}, p'_{\ell_{2}}, ..., p'_{\ell_{K}}\}, D_{\ell}, \{q'_{\ell_{0}}, q'_{\ell_{1}}, ..., q'_{\ell_{M}}\} \right)\; |\;\text{ if } p_{\ell_{m+1}} \neq 0 \text{ then } p'_{\ell_{m}} = p_{\ell_{m+1}} + \frac{p_{\ell_1}}{|{p_{nz}} - 1|}, \\ & & \text{ otherwise } p'_{\ell_m} = p_{\ell_{m+1}},
    D_{\ell}' = D_{\ell} - 1, q'_{\ell_n} = q_{\ell_{n+1}}, q'_M = 0 \; \text{ for } m = 0,...,K-1 \text { and }\\ & & n =  0,...,M-1, \\
     r'_i &=& (\{p'_0, p'_1, ..., p'_K \}, D'_i, \{q'_0, q'_1, ..., q'_M \}), i \neq \ell |\; p'_i = p_i, D'_i = D_i - 1, q'_m = q_{m+1}, q_M' = 0 \;,\\ & & m = 0,...,M-1, 
  \end{eqnarray*}  
  
  and $\delta(s,a,s') = p_{{\ell}_1}$ if:
  \begin{eqnarray*}
    \exists r_i &=& (\{p_{i_{0}}, p_{i_{1}}, p_{i_{2}}, ..., p_{i_{K}}\}, D_i, \{q_{i_{0}}, q_{i_{1}}, ..., q_{i_{M}}\}) \in s | p_{i_1} \neq 0, D_i \neq 0, q_{i_1} = 0, \\
    a &=& \ell, \\
    r_{\ell}' &=& (\{p'_{\ell_{0}}, p'_{\ell_{1}}, p'_{\ell_{2}}, ..., p'_{\ell_{K}}\}, D_{\ell}, \{q'_{\ell_{0}}, q'_{\ell_{1}}, ..., q'_{\ell_{M}}\})\; |\; p'_{\ell_0} = 1, p'_{\ell_m} = 0, D'_{\ell} = D_{\ell} - 1, \\ && q'_{\ell_n} = q_{\ell_{n+1}}, q_M' = 0 \text{ for } m = 1, ..., K \text{ and } n = 0,...,M-1, \\
    r_i' &=& ((\{p'_0, p'_1,..., p'_K\}, D'_i, \{q_0, q'_1, ..., q'_M\})\;, i \neq \ell |\; \mathbf{p}'_i = \mathbf{p}_i, D_i' = D_i - 1, q'_{i_n} = q_{i_{n+1}}, q_M' = 0, \; \\ & & \text{ for } n = 0,...,M-1.
\end{eqnarray*}    
  

\begin{example}
In the MDP presented in Figure \ref{fig:MDPExample}, this case occurs when transitioning from S3 to S4 or S5. When the agent chooses action 1, the agent could transition to S5 with probability 0.5. In S5, the computation time becomes 0 for the $1^{st}$ request with probability 1 and all other probabilities in the distribution become 0, as the task has completed (e.g., $ \{p_{1_0} = 0, p_{1_1} = 0.5, p_{1_2} = 0.5, p_{1_3} = 0, p_{1_4} = 0\} \rightarrow \{p'_{1_0} = 1, p'_{1_1} = 0, p'_{1_2} = 0, p'_{1_3} = 0, p_{1_4} = 0\} $). On the other hand, the agent can also transition to S4 with probability 0.5. In S4, the computation time for the $1^{st}$ request becomes 1 with probability 1 (e.g, $ \{p_{1_0} = 0, p_{1_1} = 0.5, p_{1_2} = 0.5, p_{1_3} = 0, p_{1_4} = 0\} \rightarrow \{p'_{1_0} = 0, p'_{1_1} = 1, p'_{1_2} = 0, p'_{1_3} = 0, p'_{1_4} = 0\}$ as $p'_{l_{1}} = p_{l_{2}} + \frac{p_{l_1}}{ |p_{nz}-1|} = 0.5 +\frac{0.5}{(2-1)} = 1$ ).  In both states, the deadline ($ D_1 = 7 \rightarrow D_1' = 6 $) and inter-arrival time decrement (e.g., for request 1, $ \{q_{1_0} = 0, q_{1_1} = 0,... q_{1_6} = 1, q_{1_7} = 0, q_{1_8} = 0\} \rightarrow \{q_{1_0}' = 0, q_{1_1}' = 0,..., q_{1_5}' = 1, q_{1_6}' = 0, q_{1_7}' = 0, q_{1_8}' = 0\}$).
\end{example}
  
{\bf Case 4: Agent chooses to schedule a completed request ($\mathbf{ a = \ell \wedge p_0 = 1 \wedge q_1 \neq 1)}$).} When the completion time for a request $r_{\ell}$ in the system equals 0 with a probability of 1  ($p_0 = 1$) and the agent chooses to work on the request ($A = \ell$), the action $A = \ell$ will be treated like the $None$ action (i.e., {\bf Case 1}).  This causes the inter-arrival times of requests and all non-zero deadlines to decrement. The completion time of all requests remains the same. Deadlines of soft requests which are zero remain zero, and if a deadline of a hard request is zero, see {\bf Case 6}. Formally, $\delta(s,a,s') = 1$ if: 
\begin{eqnarray*}
    \exists r_{\ell} &=& (\{p_0, p_1, ..., p_K \}, D_{\ell}, \{q_0, q_1, ..., q_M \}) \in s| p_{0} = 1, q_{1} = 0,\\
  a &=& \ell,\\
  \forall r_i' &=& (\{p_0', p_1', ..., p_K' \}, D_i', \{q_0', q_1', ..., q_M' \}) \in s',\; \mathbf{p}'_i = \mathbf{p}_i,\; \text{if} \; D_i \neq 0,\; D'_i = D_i - 1, \\
  & & \text { otherwise } (D'_i = D_i = 0 \wedge r_i \in r_{soft}), q'_n = q_{n+1} \; , q'_M = 0 \text { for } n = 0,...,M-1.
\end{eqnarray*}

{\bf Case 5: Soft request(s) missed their deadline(s), and are not yet replaced ($\mathbf{ r_i \in r_{soft} \wedge p_0 \neq 1 \wedge }$ 
$\mathbf{D_i = 0 \wedge q_1 = 0}$).} In this case, the soft requests that missed their deadlines have a deadline of 0 
in their next state. 
The deadlines of all other requests and the inter-arrival times of every request will decrement. The completion time of the request selected (e.g., $a = \ell$) will also decrement.  All non-selected requests will keep their completion times the same. Formally, $\delta(s,a,s') = 1$ if:
\begin{eqnarray*}
    \exists r_i &=& (\{p_0, p_1, ..., p_K\}, D_i, \{ q_0, q_1, ..., q_M \}) \in s \;|\; D_i = 0, q_{1} = 0 \\
       a &=& \ell,\\
    r_{\ell}' &=& (\{p'_0, p'_1, ..., p'_K\}, D'_{\ell}, \{ q'_0, q'_1, ..., q'_M \}) \in s' | \; p'_m = p_{m+1}, p'_K = 0, \\ & & \text{ if } (D_{\ell} = 0 \wedge r_{\ell} \in r_{soft}) \vee (r_{\ell} \in r_{hard} \wedge p_0 = 1) \\
    & & \text{then }D_{\ell}' = 0, \text{ otherwise } D'_{\ell} = D_{\ell}-1, q_n' = q_{n+1}, q_M' = 0\text{ for } m =0,..., K-1, \\ && \text{ and } n = 0,..., M-1,\\
    r_i' &=& (\{p'_0, p'_1,..., p'_K\}, D_i', \{q_0, q'_1, ..., q'_M\}), i \neq \ell |\; \mathbf{p}'_m = \mathbf{p}_{m}, \\ && \text{ if } (D_i = 0 \wedge r_i \in r_{soft}) \vee (r_i \in r_{hard} \wedge p_0 = 1)  \text{ then }D'_i = 0, \\ 
    & & \text{ otherwise } D'_i = D_i-1, q'_n = q_{n+1} \; , q'_M = 0 \; \text{for } n = 0,..., M-1, \text{ and } \\
    R &=& -10, \forall r_i \in r_{soft} \text{ with } D'_i = 0 \wedge D_i \neq 0.
\end{eqnarray*}   
 
  
\begin{example}
Consider S23 in the MDP in Figure \ref{fig:MDPExample} where the hard task $r_1 = ( \{p_0 = 1.0, p_1 = 0, p_2 = 0, p_3 = 0, p_4 = 0\}, 0, \{q_0 = 0, q_1 = 1, q_2 = 0, q_3 = 0,..., q_9 = 0\})$ is complete and the soft task $r_2 = (\{p_0 = 0, p_1 = 1, p_2 = 0\}, 0, \{q_0 = 0, q_1 = 1, q_2 = 0, q_3 = 0, q_4 = 0\})$ has missed its deadline and incurred a cost of $R = -10$. Unlike in Figure \ref{fig:MDPExample}, if the soft request were not scheduled for replacement (e.g., if $q_1 \neq 1)$ and the soft request is worked on, the next state would complete the task $r'_2 = (\{p'_0 = 1.0, p'_1 = 0, p'_2 = 0\}, 0, \{q'_n = q_{n+1},...\})$.  Note if the completed hard task were not scheduled for replacement, it would decrement normally (e.g., $r'_1  = ( \{p'_0 = 1.0, p'_1 = 0, p'_2 = 0, p'_3 = 0, p'_4 = 0\}, 0, \{q'_n = q_{n+1},...\})$).
\end{example}


{\bf Case 6: Hard request misses deadline ($\mathbf{ D_i=0 \wedge r_i \in r_{hard} \wedge p_1 \neq 0}$).} When a request has a hard deadline $D$ and the trip is incomplete by $D$, the next state is the terminal state with $p = 1$. Formally, $\delta(s,a,s') = 1$ if:
  \begin{eqnarray*}
  \exists \; r_i &=& (\{p_0, p_1, ..., p_K\}, D_i, \{ q_0, q_1, ..., q_M \}) \in s | p_0 \neq 1, D_i = 0, r_i \in r_{hard}, \\
  \forall \; a &\in & A, \\
  s' &=& s_{term}. 
  \end{eqnarray*}
 
\begin{example}
In the MDP in Figure \ref{fig:MDPExample}, this case occurs when transitioning from S12 to S13. From S12, the deadline for $r_1$ will reach $0$ regardless of the action taken. When the agent chooses any action in $A$, it transitions to the terminal state S13 with probability 1.0.
\end{example}

{\bf Case 7: New instance of request ${\mathbf r_{\ell}}$ arrives with probability $\mathbf{q_{\ell_{1}}}$.}
When the inter-arrival time for a request $r_{\ell}$ in the system equals 1 with a non-zero probability $q_{{\ell}_1}$, the request will be replaced with a new instance with a probability $q_{\ell_1}$ in the next timestep. On the other hand, the request will persist with probability $(1 - q_{\ell_1})$. In this case, the remaining non-zero probability entries of the next state will be increase by $q_{\ell_1} /|q_{nz}-1|$ where $|q_{nz}|$ is the number of inter-arrival times with non-zero probability. Note that multiple requests may have new instances arriving simultaneously. Formally, $\delta(s,a,s') = 1 - q_{\ell_1}$ if:
  
  \begin{eqnarray*}
  \exists r_{\ell} &=& (\{p_{\ell_{0}}, p_{\ell_{1}}, p_{\ell_{2}}, ..., p_{\ell_{K}}\}, D_{\ell}, \{q_{\ell_{0}}, q_{\ell_{1}}, ..., q_{\ell_{M}}\}) \in s| q_{\ell_{1}} \neq 0,\\
  a &=& j,\\
    r_{\ell}' &=& (\{p'_{\ell_{0}}, p'_{\ell_{1}}, p'_{\ell_{2}}, ..., p'_{\ell_{K}}\}, D'_{\ell}, \{q'_{\ell_{0}}, q'_{\ell_{1}}, ..., q'_{\ell_{M}}\})\; |\;\text{ if } (\ell= j) \wedge (p_0 \neq 1),p'_{\ell_{m}} = p_{\ell_{m+1}}, p_{\ell_K} = 0, \\ & & \text{ otherwise } \mathbf{p}'_{\ell} = \mathbf{p}_{\ell},  
   \text { if } D_{\ell} \neq  0,\; D'_{\ell} = D_{\ell} -1, \text { otherwise } D'_{\ell} = 0, \text{ if } q_{\ell_{n+1}} \neq 0 \\ & & \text{ then } q'_{\ell_{n}} = q_{\ell_{n+1}} + \frac{q_{\ell_1}}{|{q_{nz}} - 1|}, \\ & &\text{ otherwise } q'_{\ell_n} = q_{\ell_{n+1}}, q'_{\ell_M} = 0, \text{ for } m = 0,...,K-1, \text{ and }n = 1,..., M-1. 
  \end{eqnarray*}
  
\begin{example}
In the MDP presented in Figure \ref{fig:MDPExample}, this case can be seen when transitioning from S4 or S5 to S6. When the agent chooses any action, the agent transitions to S6. In S6, the inter-arrival time becomes 0 for the $1^{st}$ request. This results in the $1^{st}$ request being replaced by a new request in its initial configuration $(\{p_0 = 0, p_1 = 0, p_2 = 1\}, 0, \{q_0 = 0, q_1 = 1,... q_4 = 0\})  \rightarrow (\{p_0 = 0, p_1 = 0, p_2 = 1\}, 3, \{q_0 = 0, q_1 = 0,... q_4 = 1\})$.  
\end{example}
\begin{example}
An example of two new requests arriving simultaneously occurs in states S9, S20, and S23. Under any action chosen in those three states, a hard request and a soft request will arrive. The initial configuration of S1 is then recovered $[(\{p_0 = 0, p_1 = 0, p_2 = 1\}, 0, \{q_0 = 0, q_1 = 1,... q_4 = 0\}) , (\{p_0 = 1, p_1 = 0, p_2 = 0, p_3 = 0, p_4 = 0\}, 0, \{q_0 = 0, q_1 = 1,... q_8 = 0\})] \rightarrow [( \{p_0 = 0, p_1 = 0, p_2 = 1\}, 3, \{q_0 = 0, q_1 = 0,... q_4 = 1\}) , ( \{p_0 = 0, p_1 = 0, p_2 = 0, p_3 = 0.5, p_4 = 0.5\}, 7, \{q_0 = 0, q_1 = 0,... q_8 = 1\})]$.  This is shown in the loops from S9, S20, and S23 to S1.
\end{example}

{\bf Case 8: Agent in terminal state (s = s_{term}).}
In this case, the next state will also be the terminal state regardless of action taken. Formally, $\delta(s,a,s') = 1$ if:
\begin{eqnarray*}
s &=& s_{term} \\
\forall \; a &\in & A \\
s' &=& s_{term}
\end{eqnarray*}

\begin{example}
In the MDP in Figure \ref{fig:MDPExample}, this case occurs when transitioning from S13. When the agent chooses any action from S13, the agent transitions back to S13 with probability 1.0. 
\end{example}


\subsection{Preemptible vs. Non-Preemptible Scheduling}
\label{subsec:PrevNonPre} 
Two types of MDPs are defined. A \emph{preemptible} MDP is an MDP without any constraints on actions it can take at any given state. A request can be stopped before it is completed and another request can be started without any consequences.  Unlike a preemptible MDP, a \emph{non-preemptible} MDP may have a limited set of actions it can take at any given state. Specifically, an agent traversing a non-preemptible MDP must finish a request before starting a new request, and no idle time can be inserted into the completion of the request (e.g., $a \neq None$). If a new instance of a request arrives before an agent completes the current request, the agent will have the option to start a different request instead of working on the new instance.


\subsubsection{Non-Preemptible MDP State Space}
\label{subsubsec:PrevNonPre}

The state space for non-preemptible MDPs can be partitioned into two subsets: (1) Restricted states ($S_{r}$), in which an agent only has access to one action and (2) Unrestricted states ($S_{u}$), in which an agent has access to more than one action.  The implementation of the non-premptible MDP abstracts restricted states from the preemptible MDP, resulting in a smaller state-space (e.g., successor states resulting from the repeated application of a single action are abstracted into a single state).

\begin{example}
In the non-preemptible MDP based on the MDP presented in Figure \ref{fig:MDPExample}, S1 would be considered an unrestricted state since the traversing agent is free to choose any action $a \in A$. However, once the agent transitions to S2 by choosing action 1, it is unable to choose any request but 1 until $r_1$ is complete. As a result, S2 would be considered a restricted state. In this implementation S2, S3, and S4 would be abstracted as restricted states and the agent would transition directly from S1 to S5 or S6 after choosing action 1.
\end{example}

\section{Generating the Safe Scheduler}
\label{Sec:SafeSch} 
\begin{figure} [h!]
        \centering
        \includegraphics [width=15cm, height=15cm, keepaspectratio]
        {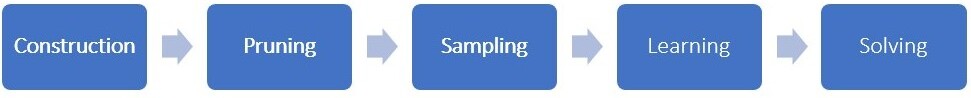}
        \caption{Process to solve the UAM Scheduling Problem}
        \label{fig:flow}
\end{figure}

 The high-level process for solving a UAM scheduling problem is shown in Figure \ref{fig:flow}. The process begins by constructing an MDP given a set of routes and requests as described in Subsection \ref{subsec:MDPC} (Construction). The MDP is then pruned to remove terminal cases so agents can safely traverse the MDP while sampling (Pruning). The agent traverses the pruned MDP to sample the completion and inter-arrival time distributions (Sampling). After sampling the distributions, the agent can create an estimate MDP (Learning), which it uses to find the optimal policy by employing value iteration or MCTS (Solving). \\ 
 
\vspace{-20 pt}        
\subsection{Pruning the MDP}
\label{subsec:Prune}

The scheduler begins by pruning the MDP to remove any terminal states from the system. A backwards reachability algorithm is used to prune states and actions that have non-zero probability of leading to the terminal state. If all actions for a given state are pruned, then the state is also pruned from the MDP, and that state is labelled as a terminal state.  The same pruning step is then performed for all actions and states which lead to the newly labelled terminal state. For example, in the MDP figure diagram, the agent will begin by pruning the Terminal state ($S13$). From the terminal state, the algorithm will backtrace to state $S12$ and prune the actions $\{1, 2, None\}$, as they all lead to the Terminal state. Since all the actions at $S12$ are pruned, the algorithm will prune $S12$ from the system and repeat the process for all actions and states which lead to $S12$. As a result, the algorithm will prune any intermediate states between $S11$ and $S12$ as they inevitably lead to the Terminal state. At $S11$, the pruning step will remove the None and $2$ action as choosing any action other than $1$ will lead to a pruned state and, by extension, the Terminal State.  This process applies to both a preemptible and non-preemptible MDP.


\subsection{Sampling the MDP}
\label{subsec:SA}   
After pruning, the scheduling agent begins traversing the pruned MDP. When sampling a specific request, the agent will choose to work on the specific request at all time steps except when prevented by the pruning step. When the request is completed, the agent increments a counter corresponding to the number of time steps required to finish the request. The agent repeats this process for a given number of samples. Once the specified sample number is reached, the agent normalizes the counters across the number of samples to obtain an estimated probability distribution. The agent performs the same process when estimating the inter-arrival time distribution and records when a new instance of the same request appears. The agent repeats the sampling process for all requests in the system to have an accurate estimate of the completion and inter-arrival times of all requests. The number of samples is determined by the formula outlined in \cite{GGR2018}. Once all requests have been sampled, the agent uses the estimated distributions to create an estimate request system and an estimate MDP. By running value iteration or MCTS on the estimate MDP, the scheduling agent can choose the near-optimal action at each time step.

The difference in probability distributions between the pruned and unpruned MDPs comes from paths that would lead to the terminal state, which should not be taken.  By removing those paths, the estimated MDP only captures the safe traces of the actual MDP, and thus the MDP is safely learned.

   \begin{figure}[h!]
    \centering
    \includegraphics [width=10cm, height=10cm, keepaspectratio] {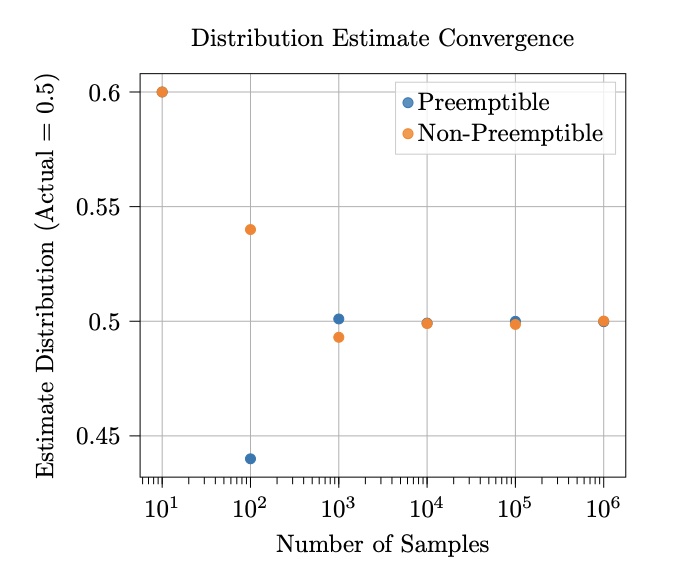}
    \caption{Delay Distribution for Priority Route with Completion Time=\{1,2\}}
    \label{fig:DelayDis}
    \end{figure}

The accuracy of the sampling algorithm is shown in Figure \ref{fig:DelayDis}. A preset MDP was used as a plant for experimentation to gauge how well the agent could estimate the MDP. In [4], the authors derive that $y = r * \left \lceil{\frac{1}{2 \epsilon^2} (\ln{2r} - \ln{\gamma} )}\right \rceil  $ in order to establish the relationship between $\epsilon$, the percentage to which the estimated MDP converges to the actual MDP, and y, the numbers of samples taken over the plant. In this expression, r represents the size of the support of the estimated distribution and $1 - \gamma$ is the probability of the estimated model being within $\epsilon$ of the actual model. A value of 1000 samples is taken in this work, with $1 - \gamma = 0.9$ and $r = 2$, which yields epsilon as approximately 0.0607.  The estimated MDP is within 6.07 \% of the actual MDP and has approximately 94\% accuracy. The completion time distribution of $\{q_0 = 0, q_1 = 0.5, q_2 = 0.5\}$ was the configuration being estimated and a set number of samples ($1000$) was employed. As the number of samples increased, the scheduler converged on the correct distribution in the preemptible and non-preemptible implementations. The MDP can be estimated using these approximations to the inter-arrival and completion time distributions. 
\subsection{Learning the MDP}
\label{subsec:LearningMPD} 
After sampling the MDP,  estimate probability distributions for computation time and inter-arrival time are created. The calculation for the estimate distributions is: $\text{\# samples for a given time}/\text{total samples}$. For example, if a sampling agent took 10 samples from a request in a system and recorded that 6 of those samples took 3 time steps to complete, it would estimate the completion time of 3 of the request as: $p_3 \in T_{est} = 6/10 = 0.6$. After creating estimates for the computation and inter-arrival times for all requests, the agent creates the estimate request set: $\{\hat{r}_{{1}}, \hat{r}_{{2}}, ..., \hat{r}_{{n}}\} = \{(\mathbf{\hat{p}}_{{1}}, D_1, \mathbf{\hat{q}}_{{1}}), (\mathbf{\hat{p}}_{{2}}, D_2, \mathbf{\hat{q}}_{{2}}),..., (\mathbf{\hat{p}}_{{n}}, D_n, \mathbf{\hat{q}}_{{n}})\}$. After estimating the request set, the construction process explained in Section \ref{subsec:MDPC} is used to create an estimated MDP. 
Since the agent cannot miss hard deadlines during sampling, it cannot gain a comprehensive estimate of the reward function. Thus, the reward function values are provided prior to sampling.  

\section{Solving for the Optimal Policy of the Learned MDP}
\label{sec:SolnMDP}
         
After estimating the MDP model, the next step is to find the optimal reward while traversing the MDP.  A policy $\pi(s)\rightarrow a$ is defined as mapping a state $s$ to an action $a$, and the goal is to find the optimal policy that maximizes (or minimizes) an objective function over the MDP's time horizon, $\tau$.  An optimal policy $\pi$ for an infinite time horizon MDP can maximize the expected discounted total reward as follows:
\begin{equation}
\label{eqn2}
\max V_{\pi}(s) = \max \left\{ \underset{\tau \to \infty}\lim  E_{\pi,s}\left[\sum_{t=0}^{\tau}\gamma_{t}R(s^{\prime}_{t}|s_t,\pi(a_t))\right] \right\}
\end{equation}
where $\gamma_t$ is the discount factor at time $t$, $s_t$ and $a_t$ are the state and action taken at decision time $t$, and $E[\cdot]$ is the expectation function. This is a stationary policy for the infinite-horizon discounted reward MDP due to the stationary nature of the transition probabilites and (bounded) rewards. The maximal achievable reward $v^{*}(s)$ at every state $s$ is found recursively by solving Bellman's equation:

\begin{equation}
\label{eqn4}
v_t^{*}(s) = \underset{a \in A}\max \left[ R_t(s,a) + \sum_{s^{\prime} \in S} P_t(s^{\prime}_{t}|s_t,a)v^{*}_{t+1}(s^{\prime}) \right]
\end{equation}

This can be done using several means: (1) Value Iteration (VI), which is optimal but computationally intensive and (2) Monte Carlo Tree Search (MCTS), which approaches optimality for large sample sizes.  Standard treatments and implementations for solving Bellman's equation using VI can be found in \cite{Put94}, \cite{Gos97}.  The MCTS approach is explained below.






            
\subsection{MCTS}
\label{subsec:SlnMCTS}
         
Unlike VI, which finds the optimal action for every state within an MDP, MCTS returns the optimal action for a given input state. The algorithm begins at the input state and deterministically takes one of the available actions. The functional steps used by the MCTS algorithm are outlined in \cite{BCGPR2021}. 
Since the MCTS algorithm’s primary input parameter is the search depth and number of samples, its runtime is independent of state space size. Thus, the scheduling agent can choose actions with a consistent run time across request configurations. 

The UAM MCTS algorithm uses Earliest-Deadline First (EDF) as the MCTS simulation policy. The EDF policy deterministically returns an action at each state in the MDP. 
When possible, the EDF policy chooses the hard request with the shortest deadline. If all hard requests have been completed, the policy chooses the soft request with the shortest deadline. This policy has been shown to be optimal when scheduling processor utilization for all tasks is less than $1$ \cite{BHR1990},\cite{CC1989}. The EDF policy is only optimal when the tasks are preemptible and a feasible schedule exists, i.e., a schedule that does not violate the deadline of a hard request \cite{CC1989}. By using EDF as the MCTS simulation policy, the scheduling agent can obtain a conditionally-optimal baseline for reward estimation at a given state and action.

\subsection{Solving the MDP:  VI vs MCTS}
\label{subsec:SlnMDPVI}
            
VI and MCTS were used to approximate the optimal policy over the learned MDP.  The route configuration seen in Fig. \ref{fig:MDPExample} for both the preemptible and non-preemptible scheduling MDP implementation were evaluated. The scheduling agent ran through ten traversals of each MDP while choosing optimal actions. To account for probabilistic variation, 1000 trials were conducted, with each trial recording the total reward across 10 traversals, with an average being taken every 50 trials.

\begin{figure}
\centering
\begin{subfigure}{.5\textwidth}
  \centering
     \begin{tikzpicture}[scale=0.9]
    \definecolor{darkorange25512714}{RGB}{255,127,14}
    \definecolor{forestgreen4416044}{RGB}{44,160,44}
    \definecolor{lightgray204}{RGB}{204,204,204}
    \definecolor{steelblue31119180}{RGB}{31,119,180}
    \definecolor{red}{RGB}{180,100,100}
    \definecolor{blue1}{RGB}{255, 0, 0}
    \definecolor{blue2}{RGB}{64, 224, 208}
    \definecolor{blue3}{RGB}{4, 55, 242}
    \definecolor{lightgray204}{RGB}{204,204,204}
    \definecolor{red1}{RGB}{164, 42, 4}
    \definecolor{red2}{RGB}{227, 11, 92}
    \definecolor{red3}{RGB}{112, 41, 99}
    \begin{axis}[
    legend style={
      fill opacity=0.8,
      draw opacity=1,
      text opacity=1,
      at={(0,0.5)},
      anchor=west,
      draw=lightgray204
    },
    tick align=outside,
    tick pos=left,
    title={Cost Incurred Over Trials},
    x grid style={lightgray},
    xlabel={Trial},
    xmajorgrids,
    xmin=2.5, xmax=1047.5,
    xtick style={color=black},
    y grid style={lightgray},
    ylabel={Cost},
    ymajorgrids,
    ymin=-66, ymax=-45.2575,
    ytick style={color=black}
    ]
    \addplot [semithick, blue1]
    table {%
    50 -47.2
    100 -45.8
    150 -46.13333333
    200 -45.35
    250 -45.84
    300 -46.3
    350 -45.74285714
    400 -46.15
    450 -46.48888889
    500 -46.56
    550 -46.38181818
    600 -46.46666667
    650 -46.43076923
    700 -46.22857143
    750 -46.34666667
    800 -46.2375
    850 -46.27058824
    900 -46.32222222
    950 -46.48421053
    1000 -46.46
    };
    \addplot [semithick, blue2]
    table {%
    50 -65.6
    100 -64.2
    150 -61.9333333
    200 -63.1
    250 -63.84
    300 -63.5
    350 -63.8
    400 -64.325
    450 -64.33333
    500 -64.4
    550 -64.78
    600 -64.933333
    650 -65.26153
    750 -64.91428
    800 -64.96
    700 -65.175
    850 -65.2
    900 -65.0941
    950  -64.905
    1000 -64.49
    };
    \addplot [semithick, red1]
    table {%
    50 -47.8
    100 -50
    150 -50.2
    200 -50.2
    250 -49.88
    300 -50.33333333
    350 -50.74285714
    400 -50.775
    450 -50.51111111
    500 -50.54
    550 -50.25454545
    600 -50.11666667
    650 -50.06153846
    700 -50.34285714
    750 -50.44
    800 -50.45
    850 -50.34117647
    900 -50.33333333
    950 -50.33684211
    1000 -50.25
    };
    \addplot [semithick, red2]
    table {%
    50 -53.4
    100 -52.2
    150 -52.06666667
    200 -52.35
    250 -52.44
    300 -52.73333333
    350 -52.51428571
    400 -52.575
    450 -52.8
    500 -53.14
    550 -52.89090909
    600 -52.9
    650 -53.13846154
    700 -53.11428571
    750 -53.12
    800 -53.375
    850 -53.30588235
    900 -53.25555556
    950 -53.33684211
    1000 -53.3
    };
    \end{axis}
    \end{tikzpicture}
  \caption{Comparison of value iteration optimal policy\\ and MCTS approximation (zoom of Fig \ref{subfig:MCTSRan} y-axis)}
  \label{subfig:PvNPAvRe}
\end{subfigure}%
\begin{subfigure}{.5\textwidth}
  \centering
    \begin{tikzpicture} [scale=0.90]
    \definecolor{blue1}{RGB}{255, 0, 0}
    \definecolor{blue2}{RGB}{64, 224, 208}
    \definecolor{blue3}{RGB}{4, 55, 242}
    \definecolor{lightgray204}{RGB}{204,204,204}
    \definecolor{red1}{RGB}{164, 42, 4}
    \definecolor{red2}{RGB}{227, 11, 92}
    \definecolor{red3}{RGB}{112, 41, 99}
    
    \begin{axis}[
    legend cell align={left},
    legend style={
      fill opacity=0.8,
      draw opacity=1,
      text opacity=1,
      at={(0.26,0.33)},
      anchor=west,
      draw=lightgray204
    },
    tick align=outside,
    tick pos= left,
    title={Cost Incurred Over Trials},
    unbounded coords=jump,
    x grid style={lightgray},
    xlabel={Trial},
    xmajorgrids,
    xmin=2.5, xmax=1047.5,
    xtick style={color=black},
    y grid style={lightgray},
    ylabel={Cost},
    ymajorgrids,
    ymin=-125, ymax=-45.0375,
    ytick style={color=black}
    ]
    \addplot [semithick, blue1]
    table {%
    50 -47.2
    100 -45.8
    150 -46.13333333
    200 -45.35
    250 -45.84
    300 -46.3
    350 -45.74285714
    400 -46.15
    450 -46.48888889
    500 -46.56
    550 -46.38181818
    600 -46.46666667
    650 -46.43076923
    700 -46.22857143
    750 -46.34666667
    800 -46.2375
    850 -46.27058824
    900 -46.32222222
    950 -46.48421053
    1000 -46.46
    nan nan
    nan nan
    };
    \addlegendentry{PE Value Iteration}
    \addplot [semithick, blue2]
    table {%
    50 -65.6
    100 -64.2
    150 -61.9333333
    200 -63.1
    250 -63.84
    300 -63.5
    350 -63.8
    400 -64.325
    450 -64.33333
    500 -64.4
    550 -64.78
    600 -64.933333
    650 -65.26153
    750 -64.91428
    800 -64.96
    700 -65.175
    850 -65.2
    900 -65.0941
    950  -64.905
    1000 -64.49
    nan nan
    nan nan
    };
    \addlegendentry{PE MCTS(EDF)}
    \addplot [semithick, blue3]
    table {%
    50 -117.6
    100 -117.5
    150 -119.2
    200 -118.45
    250 -119.08
    300 -119.333
    350 -119.257
    400 -119.6
    450 -119.489
    500 -120.3
    550 -120.036
    600 -120.233
    650 -120.369
    700 -119.986
    750 -120.253
    800 -120.125
    850 -120.365
    900 -120.364
    950 -119.989
    1000 -119.74
    nan nan
    nan nan
    };
    \addlegendentry{PE MCTS(Random)}
    \addplot [semithick, red1]
    table {%
    50 -47.8
    100 -50
    150 -50.2
    200 -50.2
    250 -49.88
    300 -50.33333333
    350 -50.74285714
    400 -50.775
    450 -50.51111111
    500 -50.54
    550 -50.25454545
    600 -50.11666667
    650 -50.06153846
    700 -50.34285714
    750 -50.44
    800 -50.45
    850 -50.34117647
    900 -50.33333333
    950 -50.33684211
    1000 -50.25
    nan nan
    nan nan
    };
    \addlegendentry{Non-PE Value Iteration}
    \addplot [semithick, red2]
    table {%
    50 -53.4
    100 -52.2
    150 -52.06666667
    200 -52.35
    250 -52.44
    300 -52.73333333
    350 -52.51428571
    400 -52.575
    450 -52.8
    500 -53.14
    550 -52.89090909
    600 -52.9
    650 -53.13846154
    700 -53.11428571
    750 -53.12
    800 -53.375
    850 -53.30588235
    900 -53.25555556
    950 -53.33684211
    1000 -53.3
    nan nan
    nan nan
    };
    \addlegendentry{Non-PE MCTS(EDF)}
    \addplot [semithick, red3]
    table {%
    50 -75.2
    100 -77.1
    150 -77.73333333
    200 -77.45
    250 -76.6
    300 -76.7
    350 -76.6
    400 -76.57777778
    450 -76.44
    500 -76.67272727
    550 -76.71666667
    600 -76.71666667
    650 -76.71
    700 -77.2
    750 -76.42
    800 -76.57777778
    850 -76.73333333
    900 -76.45
    950 -76.41052632
    1000 -76.27
    nan nan
    nan nan
    };
    \addlegendentry{Non-PE MCTS(Random)}
    \end{axis}
    \end{tikzpicture}
  \caption{Comparison of EDF request selection strategy to random selection in MCTS approximation}
  \label{subfig:MCTSRan}
\end{subfigure}
\caption{Average Cost for Preemptible and Non-premptible UAM Scheduling for Fig. \ref{fig:MDPExample} Request System}
\vspace{-10 pt}
\label{fig:AvRew}
\end{figure}
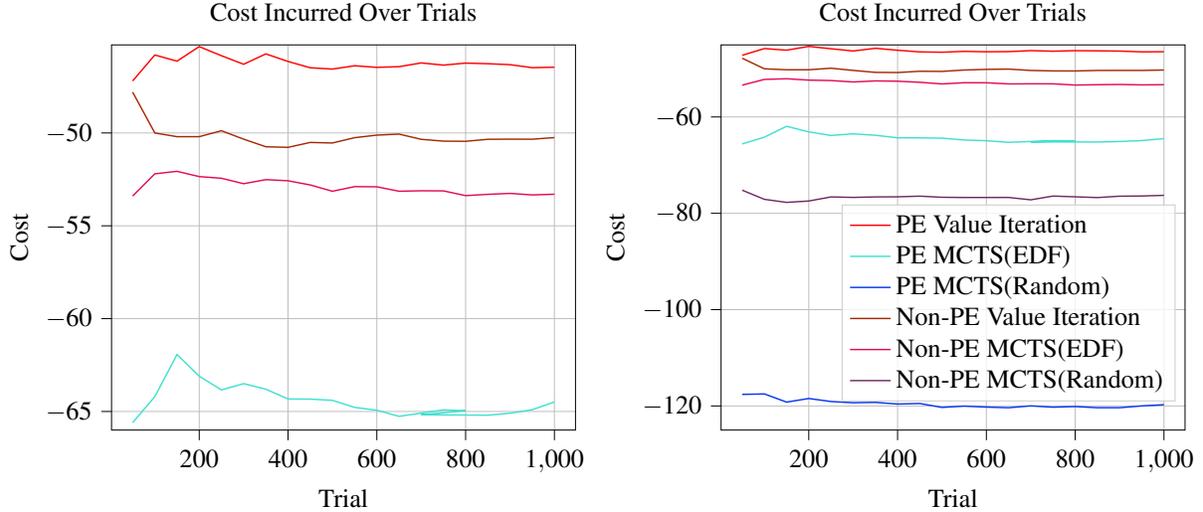

The optimal preemptible policy found through VI, as shown in Fig. \ref{subfig:PvNPAvRe}, converges to a cost of approximately $-46$ (averaged over $10$ MDP traversals and 1000 trials). The non-preemptible optimal policy obtained through VI, also shown in Fig. \ref{subfig:PvNPAvRe}, converges to a cost of about $-50$ (averaged over $10$ MDP traversals and 1000 trials). Since the non-preemptible scheduler is less flexible (i.e., has fewer states) than the preemptible, its optimal policy accrues a larger cost over time. The MCTS approximation to the optimal policy for both the preemptible and non-preemptible MDP (see Figures \ref{subfig:PvNPAvRe},\ref{subfig:MCTSRan}) is strictly inferior to the one obtained by VI, but it forms a near approximation. Increasing the depth and number of MCTS samples has that policy approach the optimal policy of VI, which is observed in the convergence of the MCTS average cost to that of VI.    

\vspace{-5 pt}
\subsection{Near-Optimality of Average Cost: MCTS Selection Strategy}
\label{subsec:AvReward}

    
The strategy used to select the next action (i.e., request) to work on strongly affects the ability of the MCTS algorithm to approximate the optimal policy obtained through VI.     
The use of random action selection for MCTS, shown in Fig. \ref{subfig:MCTSRan}, exemplifies how the EDF selection strategy, coupled with preventing the agent from working on completed requests in the preemptible case, allows MCTS to choose the optimal action more frequently. and The average cost of random MCTS is significantly larger than the average cost of EDF MCTS schedulers. This is expected, as the EDF selection strategy results in more accurate cost estimates than random MCTS at each timestep, resulting in optimal actions being selected more frequently. 

State space abstraction allows the MCTS non-preemptible near-optimal policy to outperform the MCTS preemptible near-optimal policy for both EDF and Random (see Fig. \ref{subfig:MCTSRan}). Since the simulation depth parameter of MCTS is kept constant, the MCTS algorithm is able to step farther into the future for the non-preemptible MDP thereby enabling optimal actions to be chosen more frequently at each time step. The preemptible scheduler has a large branching factor and several intermediate states. This makes it difficult to obtain an accurate future cost estimate using a random policy in the preemptible case. 
\section{Results and Discussion}
The scalability of the approach for both preemptible and non-preemptible MDPs is evaluated. The optimal policy is obtained  through VI, run with a convergence parameter of $0.99$, over a variety of task configurations as well as demand and delay distributions.  This is done on a Dell XPS 13 Windows Processor with an $11^{th}$ generation Intel Single Core i7 at 3.0 GHz.  The baseline request configuration is defined as two deterministic tasks: $r_1 \in r_{hard} = (\{p_0 = 0, p_1 = 0, p_2 = 0, p_3 = 1.0\}, 7, \{q_0 = 0, q_1 =0, ..., q_8 = 1\})$, $r_2 \in r_{soft} = (\{p_0 = 0, p_1 = 0, p_2 = 1\}, 3, \{q_0 = 0, q_1 =0, ..., q_4 = 1\})$.

\subsection{Scalability over Number of Routes}
    

\begin{table}[h!]
\centering
\begin{tabularx}{0.9\textwidth} { 
  | >{\centering\arraybackslash}X 
  | >{\centering\arraybackslash}X 
  | >{\centering\arraybackslash}X 
  | >{\centering\arraybackslash}X 
  | >{\centering\arraybackslash}X 
  | >{\centering\arraybackslash}X 
  | >{\centering\arraybackslash}X |}
 \hline
 \# Routes  & Preemptible Average Time (s) & \ Preemptible \# States  &  Non-Preemptible Average Time (s)  & \ Non-Preemptible \# States \\
 \hline
      1 Hard \& 1 Soft     &     1.438           &      47 & 0.212 &   18 \\
      \hline
      1 Hard \& 2 Soft     &     1033.546           &     82 & 126.102 &   23\\
      \hline
      1 Hard \& 3 Soft     &     5090.536          &     131 & 375.913 & 28 \\
\hline
\end{tabularx}
\caption{Scalability with Respect to Number of Routes for Value Iteration}
\label{tab:TasksPreemptv2}
\end{table}

The scalability of the technique with respect to increasing number of requests is explored. The baseline configuration is modified by adding additional copies of the soft request to the system, as inserting unmodified copies of the hard request can lead to unschedulable systems. The run time of VI (in seconds), the number of unrestricted states in the MDP, and the number of iterations the VI algorithm took to converge are then recorded. Table \ref{tab:TasksPreemptv2} shows a linear-like increase in the state size as the number of requests increases. Due to the abstracted restricted states, the state space of the non-preemptible MDP grows more slowly than the preemptible MDP. Since the time complexity of value iteration is $O(|S^2A|)$, the non-preempible MDP has a shorter run time than the preemptible MDP.

In the 1H \& 1S case, the MDP can be traversed without missing any deadlines. This allows the VI algorithm to converge in a few iterations since the optimal expected future reward is 0. Optimally traversing the other configurations results in a guaranteed negative reward, which leads to a larger number of iterations. The increase in the number of states and iterations leads to a larger computation time as the number of requests increased, making MCTS a necessary alternative once VI becomes intractable.

\subsection{Uncertain Trip Delay (Completion) and Demand (Inter-Arrival) Time Distributions}
\label{subsec:PDSF}

\begin{table}[h!]
\centering
\begin{tabularx}{0.9\textwidth} { 
  | >{\centering\arraybackslash}X 
  | >{\centering\arraybackslash}X 
  | >{\centering\arraybackslash}X 
  | >{\centering\arraybackslash}X 
  | >{\centering\arraybackslash}X 
  | >{\centering\arraybackslash}X|}
 \hline
 \ Completion Time (Delay) Values & $P$(Delay)  & Preemptible Average Time (s) & \ Preemptible \# States
 & Non-Preemptible Average Time (s)  & \ Non-Preemptible \# States\\ 
 \hline
  $ \{3\}$    & $p_{3} = 1$     &     1.438           &      47 & 0.212 &    18\\
  \hline
$ \{3, 4\}$      & $p_{3-4} = 0.5$     &     200.694         &     54 & 29.636 &    18\\
\hline
  $ \{1,2,3,4\}$    & $p_{1-4} = 0.25$     &     225.12         &     59 & 37.715 &    21\\
\hline
\end{tabularx}
\caption{Increasing Uncertainty over Delay Distribution (Completion Time) for Hard Requests}
\label{tab:ProbDelay}
\end{table}


Scalability with respect to the completion time distribution support (e.g., delay) is also investigated. The support size of the trip completion time (delay) probability distribution of the hard request is varied. Three different support sizes over completion time are used: (1) $\{p_0 = 0, p_1 = 0, p_2 = 0, p_3 = 1.0\}$, (2) $\{p_0 = 0, p_1 = 0, p_2 = 0, p_3 = 0.5, p_4 = 0.5\}$, and (3) $\{p_0 = 0, p_1 = 0.25, p_2 = 0.25, p_3 =0.25, p_4: 0.25\}$. Table \ref{tab:ProbDelay} shows a marked increase in MDP state space size. The increased number of possible completion times leads to a larger number of potential negative reward conditions, resulting in an increased number of iterations for VI to converge. 
 
 Increasing the range of the completion time support function to include 4 values introduces the case where the agent will continuously miss the soft request deadline with a probability of $p_{4}$. The introduction of this persistent negative reward leads to a larger number of iterations for convergence and results in a larger run time when combined with the larger state space.

    \begin{table}[h!]
\centering
\begin{tabularx}{0.9\textwidth} { 
  | >{\centering\arraybackslash}X 
  | >{\centering\arraybackslash}X 
  | >{\centering\arraybackslash}X 
  | >{\centering\arraybackslash}X 
  | >{\centering\arraybackslash}X 
  | >{\centering\arraybackslash}X|}
 \hline
 Inter-Arrival Time (Demand) Values & $P$(Demand)  & Preemptible Average Time (s) & \ Preemptible \# States  & Non-Preemptible Average Time (s)  & \ Non-Preemptible \# States\\ 
 \hline
  $ \{8\}$    & P = 1     &     1.438           &      47  & 0.212 &    18   \\
  \hline
$ \{8, 9\}$      & P = 0.5     &     101.663          &     201 & 27.635 &    75\\
\hline
  $ \{8, 9, 10, 11\}$    & P = 0.25     &     130.924          &     219 & 47.466 &    87\\
\hline
\end{tabularx}
\caption{Increasing Uncertainty over Demand Distribution (Inter-Arrival Time) for Hard Requests}
\label{tab:ProbArriv}
\end{table}
Similarly, scalability with respect to the support size of the inter-arrival time (demand) distribution is also studied. Three different inter-arrival support sizes are examined: (1) $ \{q_0 = 0, q_1 = 0,..., q_8 = 1.0\}$, (2) $ \{q_0 = 0, q_1 = 0,..., q_8 = 0.5, q_9 = 0.5\}$, and (3) $ \{q_0 = 0, q_1 = 0, ..., q_8 = 0.25, q_9 = 0.25, q_10 =0.25, q_11: 0.25\}$. Table \ref{tab:ProbArriv} shows that MDP state space increases in a polynomial fashion across the inter-arrival time probability distribution support function. Changing the inter-arrival time distribution impacts when new request instances arrive, and the introduction of new requests at different time steps leads to longer branches in the MDP that encompass the entire completion time and inter-arrival time of the new request. The completion time distribution only impacts the completion time of the requests and thus causes a smaller increase in the state space. However, there exists a way to traverse both the preemptible and non-preemptible MDPs without missing any deadlines in all three cases. This allows VI to converge in a small number of iterations resulting in smaller run times when compared to Table \ref{tab:ProbArriv}.


\section{Conclusion}
A safe, non-preemptible scheduler for a UAM route planning system was generated for both hard and soft requests.  The UAM trip request scheduling system was modelled as an MDP, and safe learning was used to estimate the demand and delay distributions for the requests.  The synthesized estimate MDP was then used to identify the near-optimal policy that minimized the average cost incurred by missed soft deadlines while guaranteeing no hard deadlines were missed using MCTS (EDF).  The approach was shown to be scalable in terms of number of routes and size of both delay and demand distributions. A comparison of the results with Value Iteration and MCTS (Random) was also performed.  Future work will center on improving the scalability of the approach with respect to increasing numbers of hard requests and applying the approach to generate schedulers from existing UAM schedule databases.  
\section{Downloads}
The code to run the example can be found at: https://github.com/suryakmurthy/UAM_scheduler.

\bibliographystyle{eptcs}
\bibliography{generic}
\end{document}